\documentclass[10pt,twocolumn,letterpaper]{article}

 \usepackage{iccv}              

\usepackage{
  adjustbox,
  algpseudocode,
  algpseudocode,
  algorithm,
  longtable,
  multirow,
  graphicx, 
  float,
  booktabs, makecell, multirow
}

\newtheorem{definition}{Definition}

\definecolor{iccvblue}{rgb}{0.21,0.49,0.74}
\usepackage[pagebackref,breaklinks,colorlinks,allcolors=iccvblue]{hyperref}

\def\paperID{8} 
\def\confName{ICCV}
\def\confYear{2025}

\title{Uncovering Anomalous Events for Marine Environmental Monitoring \\ via Visual Anomaly Detection}
\author{Laura Weihl\thanks{Corresponding author}\\
Computer Science Department,\\IT University of Copenhagen, Denmark\\
{\tt\small lawe@itu.dk}
\and
Nejc Novak\\
Anemo Robotics ApS\\
Copenhagen, Denmark\\
{\tt\small nejc@anemorobotics.com}
\and
Stefan H. Bengtson \qquad Malte Pedersen\\
Visual Analysis and Perception Laboratory, Aalborg University, Denmark\\
Pioneer Centre for Artificial Intelligence, Denmark\\
{\tt\small \{shbe, mape\}@create.aau.dk}
}

\begin{document}
\maketitle

\begin{abstract}
Underwater video monitoring is a promising strategy for assessing marine biodiversity, but the vast volume of uneventful footage makes manual inspection highly impractical. In this work, we explore the use of visual anomaly detection (VAD) based on deep neural networks to automatically identify interesting or anomalous events. We introduce AURA, the first multi-annotator benchmark dataset for underwater VAD, and evaluate four VAD models across two marine scenes. We demonstrate the importance of robust frame selection strategies to extract meaningful video segments.  Our comparison against multiple annotators reveals that VAD performance of current models varies dramatically and is highly sensitive to both the amount of training data and the variability in visual content that defines ``normal'' scenes. Our results highlight the value of soft and consensus labels and offer a practical approach for supporting scientific exploration and scalable biodiversity monitoring.
Project page: \url{https://vap.aau.dk/aura/}
\end{abstract}

\section{Introduction} 
\emph{``Curiouser and curiouser!''} cried Alice \cite{alice}. 
Much like Alice's astonishment at the wonders of an unexplored world, there is a big interest in both understanding and raising awareness about the importance of our marine ecosystems. 
An important tool to achieve these goals is to collect video data using underwater camera setups, a practice that is becoming increasingly popular as the technology matures and becomes more cost-efficient \cite{Nalmpanti2023}.

\begin{figure}[!ht]
    \centering

    \includegraphics[trim=0 0 0 0, clip, width=\linewidth]{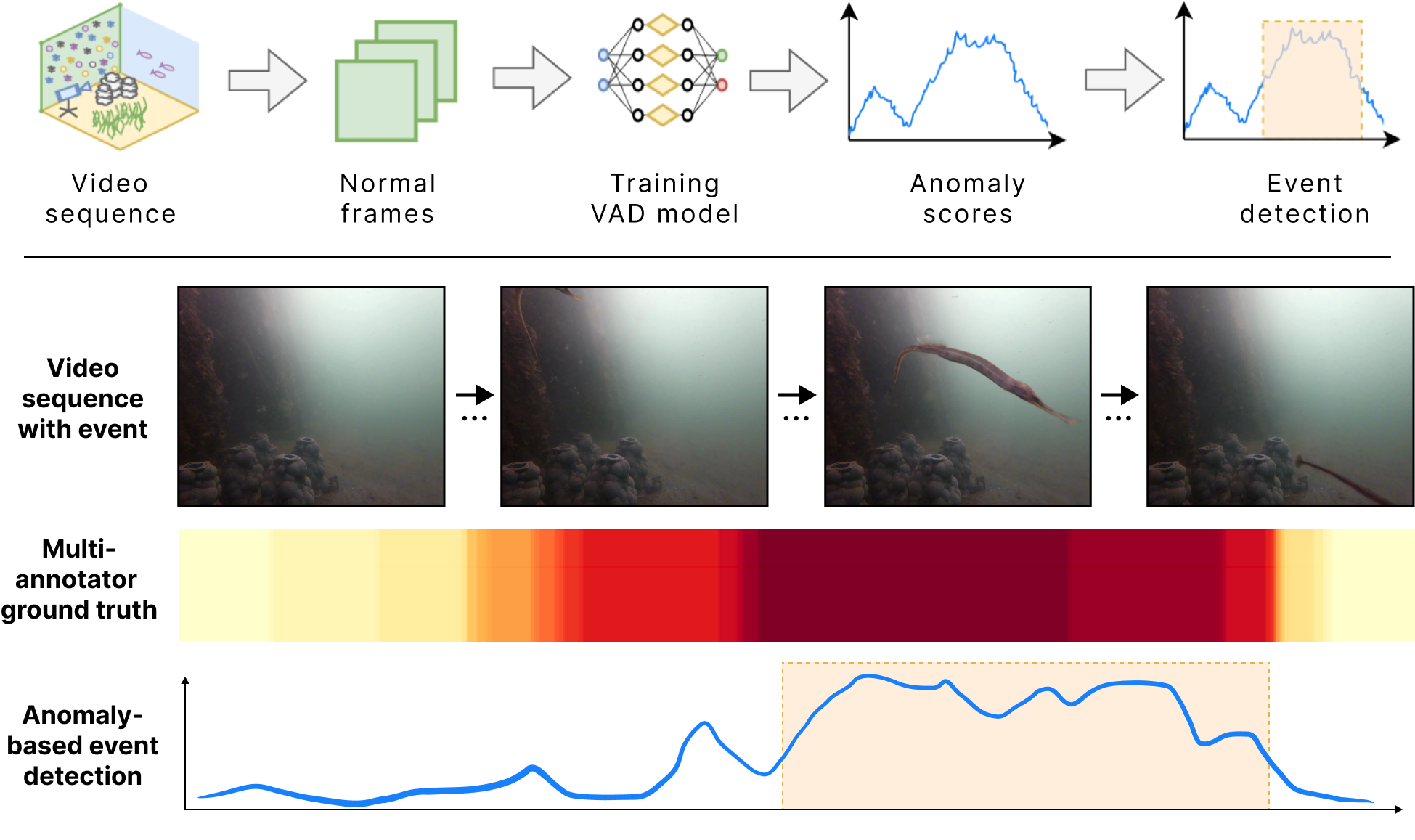}
    \caption{A VAD model trained on normal frames from an underwater camera to detect interesting events. As the fish enters the scene, the anomaly score from the model increases until the fish disappear again. The interesting event can then be detected as the sequence with a consistently high anomaly score. The multi-annotator ground truth encapsulates that some parts of the video may be less likely to be considered interesting.}

    \label{fig:intro-figure}
\end{figure}

However, a major challenge is that these video datasets are vast, with only sparse biological activity, such as occasional animal sightings or movement. 
There is hence a big need for methods to automatically extract \emph{interesting} events from such video data, which is possible using object detectors or similar models \cite{saleh2020realistic,Jansen2022FishDataset}. 
However, models trained on data from one location often fail to generalize to new locations, requiring retraining which is time-consuming and resource-intensive.

We propose addressing this challenge through visual anomaly detection (VAD), as illustrated in \cref{fig:intro-figure}.
This idea is based on the premise that interesting events can be regarded as anomalies which are rare and thus anomalous by nature, such as a fish entering an otherwise empty scene. 
Using VAD in this context is beneficial as it only requires normal data for training, i.e. empty scenes, which are easily obtainable.
Relying only on normal data also avoids having to gather examples of all possible interesting events, which can be challenging and often infeasible.
The output from the VAD, in the form of an anomaly score, can then be used to extract the interesting events from the video sequence.

VAD is often scene- and application-dependent \cite{ramachandra2020survey}. 
A small fish that is far away from the camera might not count as an anomalous occurrence but if the same fish moves closer to the camera, it could then be considered an interesting event.
What constitutes an interesting or anomalous event can therefore be highly subjective. 
Another challenging aspect of applying VAD in the underwater domain is visibility. Water turbidity and small particles of organic matter called marine snow directly impact light penetration and image quality and affect visibility under water. 
In turbid conditions, a fish might be present in a scene but is practically not visible due to the conditions.
Light levels can also vary dramatically throughout the day. 

Applying VAD in an underwater setting is therefore challenging from a computer vision perspective due to the varying visibility but also the subjective nature of defining interesting events.
We therefore also propose to rely on multiple annotators to address these issues, such that some parts of the video sequence may be associated with a lower certainty of an interesting event occurring, as also shown in \cref{fig:intro-figure}.

\noindent
The main contributions of this paper are hence as follows:

\begin{itemize}
    \item We introduce AURA (Anomalous UnderwateR Activity), the first multi-annotator dataset for visual anomaly detection in underwater scenes.
    
    \item We evaluate multiple VAD approaches on the AURA dataset, demonstrating its feasibility as a benchmark for underwater event detection, showing that VAD can uncover interesting events in marine environments.

    \item We demonstrate the necessity of using multiple annotators to account for the subjective and temporally ambiguous nature of event boundaries in dynamic underwater scenes, by evaluating how annotator differences impact model performance.
\end{itemize}



\section{Related Work}
VAD is a relatively unexplored topic in the domain of marine monitoring, the following will hence focus on VAD in general along with an overview of existing datasets dealing with marine organisms and how they could be utilized in the context of VAD.

\subsection{Visual Anomaly Detection}
VAD is a field of growing interest, which has been successfully applied in a wide variety of domains, such as industrial inspection~\cite{mvtec}, security screening~\cite{akcay2018ganomaly} and in the medical domain~\cite{Bao2024-zm}.
VAD is a concept closely related to out-of-distribution detection, however the two disciplines have two distinct objectives. In VAD, the objective is to detect the occurrence of unusual or unexpected appearance or motion. In out-of-distribution detection, the objective is to flag input samples that do not occur in the training distribution to avoid making potentially high-confidence but incorrect predictions from it~\cite{hendrycks2016baseline, goodfellow2014explaining}.
VAD is hence a good fit in the context of marine life monitoring because the collected data could contain types of life which are a rare occurrence and therefore not anticipated.
Many of the existing VAD methods are also designed to be trained solely using normal data samples~\cite{batzner2024efficientad, deng2022anomaly}, as it is often infeasible to collect a representative set of anomalies due to their rare nature.
This is also a huge benefit in the context of marine life monitoring, where normal data will often be available in great quantities. 

Common approaches for VAD are reconstruction-based methods, which are based on the idea of training the model with only normal images.
This causes the trained model to fail the reconstruction when presented with anomalous samples during inference and thereby indicating the presence of an anomaly.
An encoder-decoder architecture is widely used for these reconstruction-based methods, which in the simplest form consists of a Convolutional Autoencoder \cite{mvtec}.
Several VAD approaches expand on this encoder-decoder architecture, either through knowledge distillation in the form of a student-teacher framework~\cite{deng2022anomaly,batzner2024efficientad} or by combining it with a generative adversarial network (GAN)~\cite{akcay2018ganomaly}.
Other approaches rely solely on knowledge distillation in a student-teacher framework~\cite{wang2021student} to achieve a reconstruction-based VAD approach.
Another group of VAD methods is based on an embedding-based approach, where a similarity metric is used to detect anomalies.
The embeddings could be created from a pretrained convolutional neural network~\cite{Defard2020PaDiM, Roth2022PatchCore} and a nearest neighbor search based on these embeddings are then used to determine whether a sample is considered anomalous or not~\cite{Roth2022PatchCore}.
In this work we will primarily focus on reconstruction-based VAD methods.



\begin{figure*}[!t]
    \centering
    \includegraphics[width=0.30\linewidth]{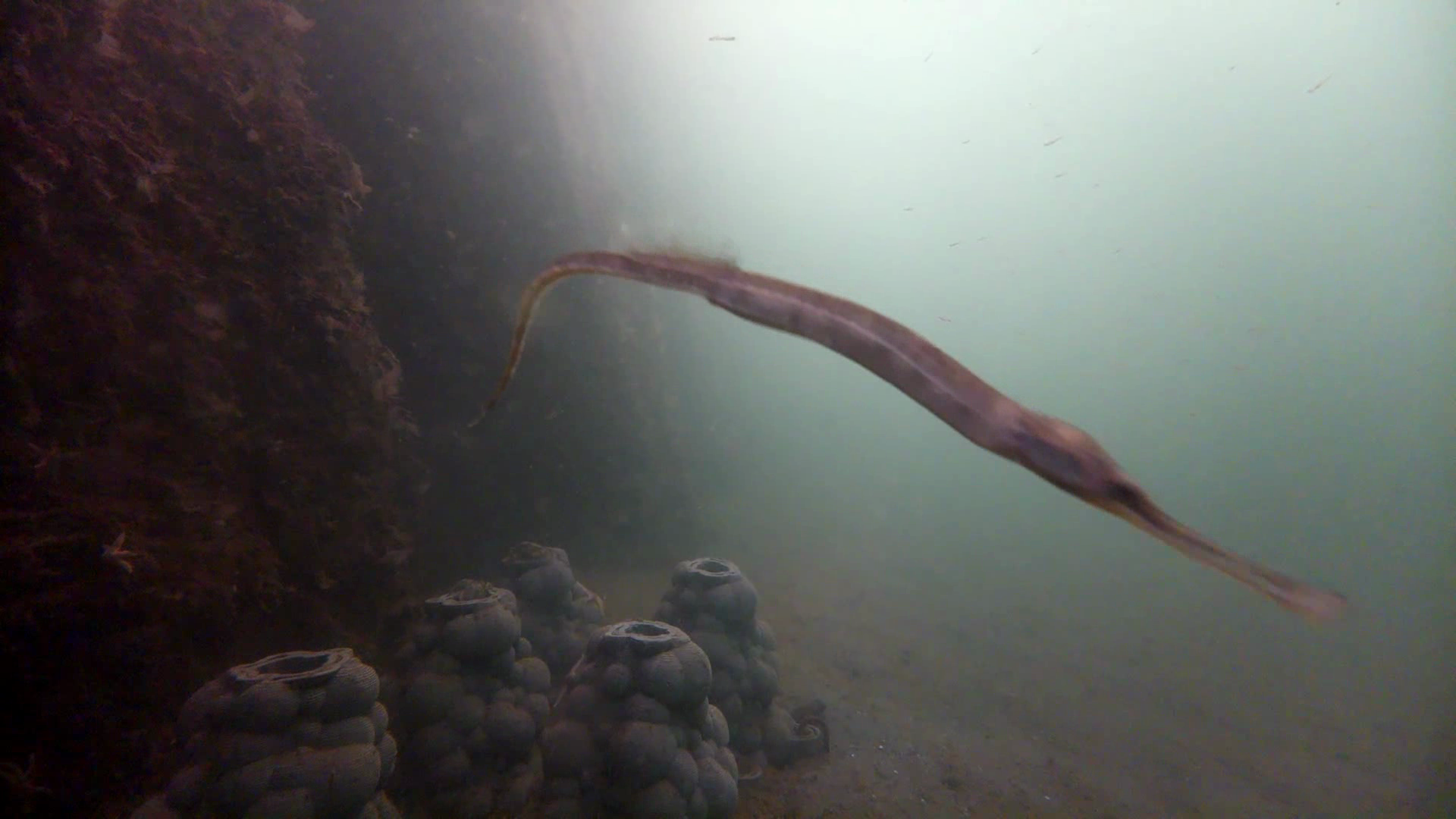}
    \includegraphics[width=0.30\linewidth]{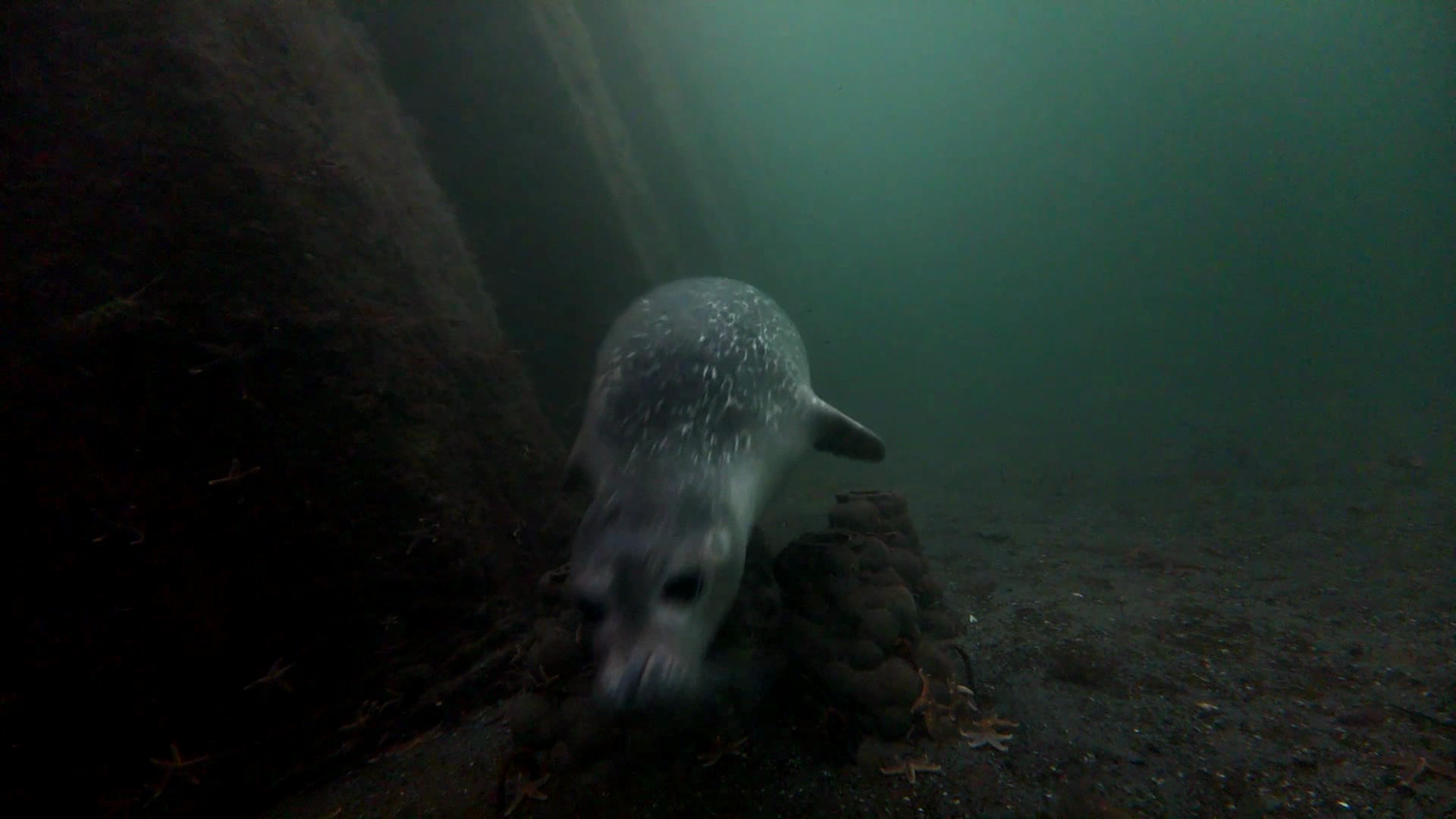}
    \includegraphics[width=0.30\linewidth]{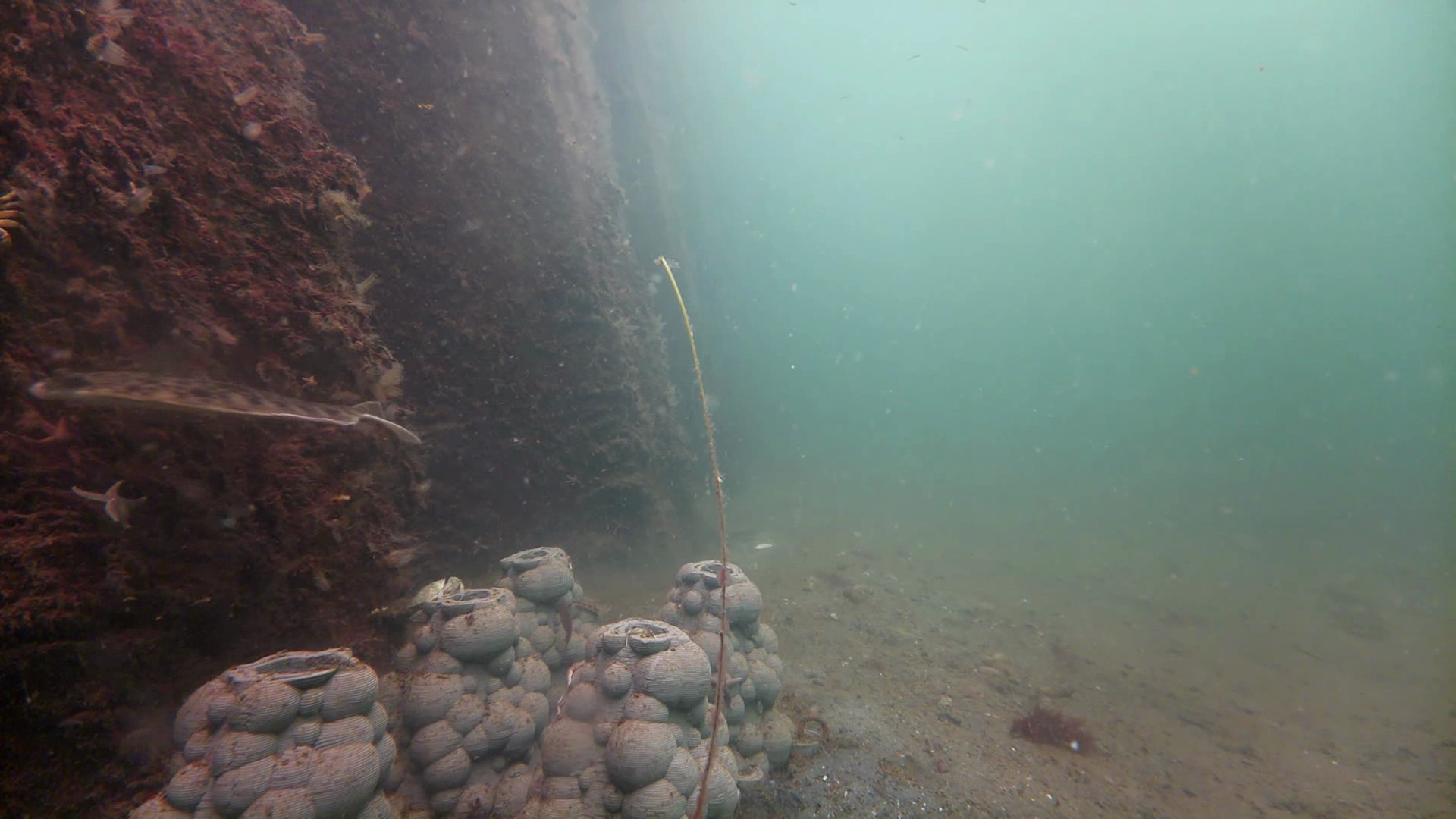}
    \includegraphics[width=0.30\linewidth]{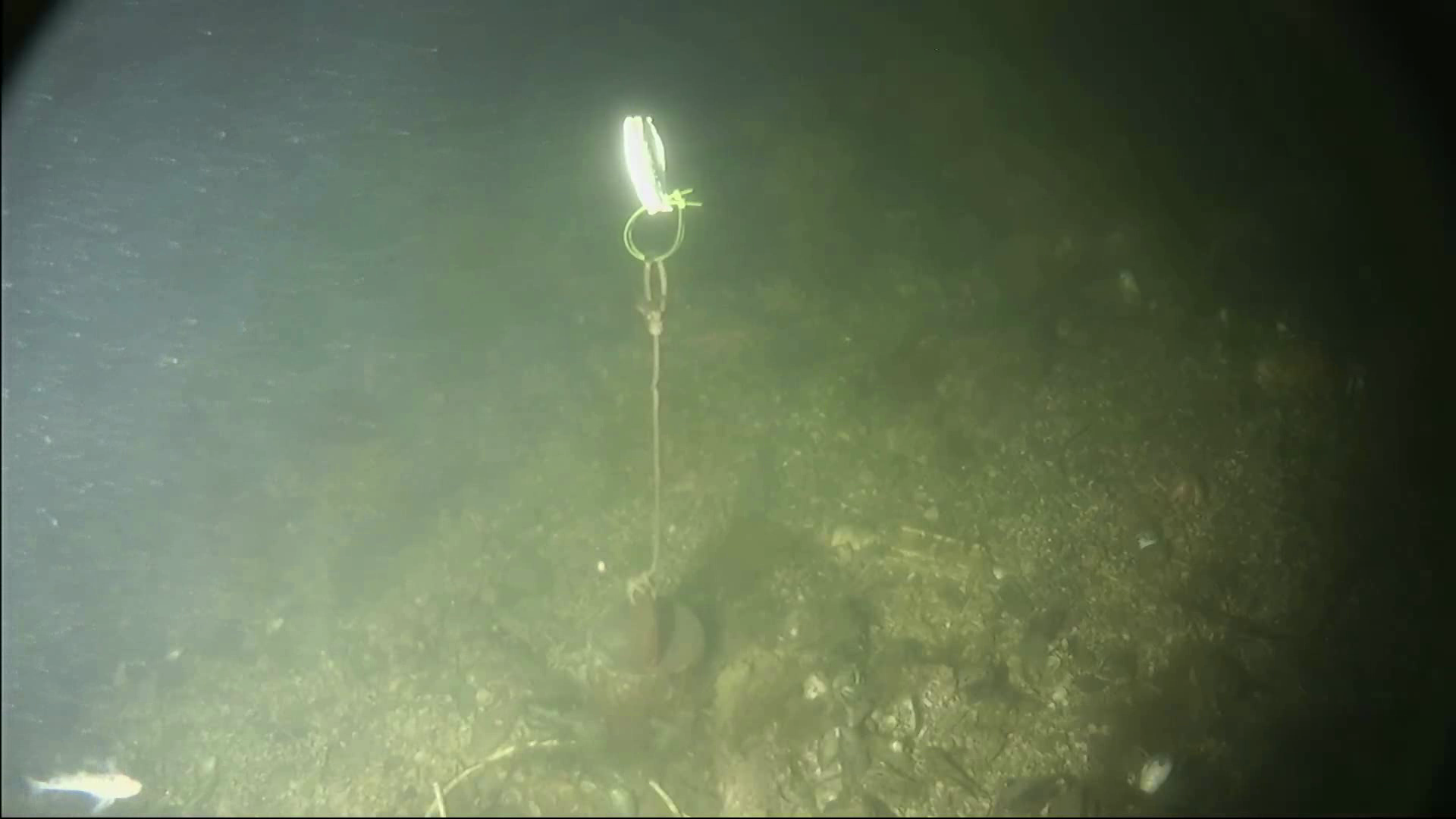}
    \includegraphics[width=0.30\linewidth]{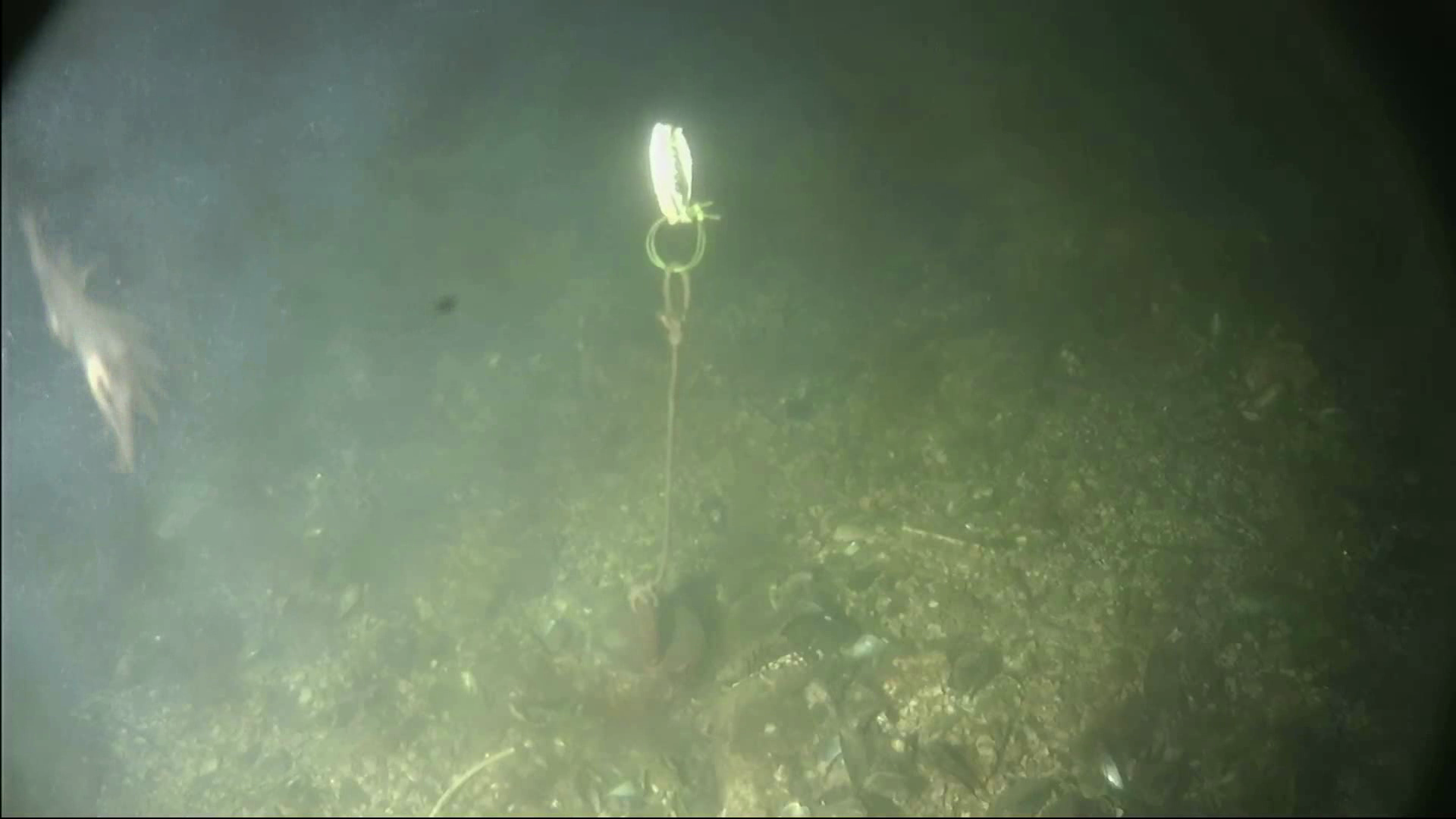}
    \includegraphics[width=0.30\linewidth]{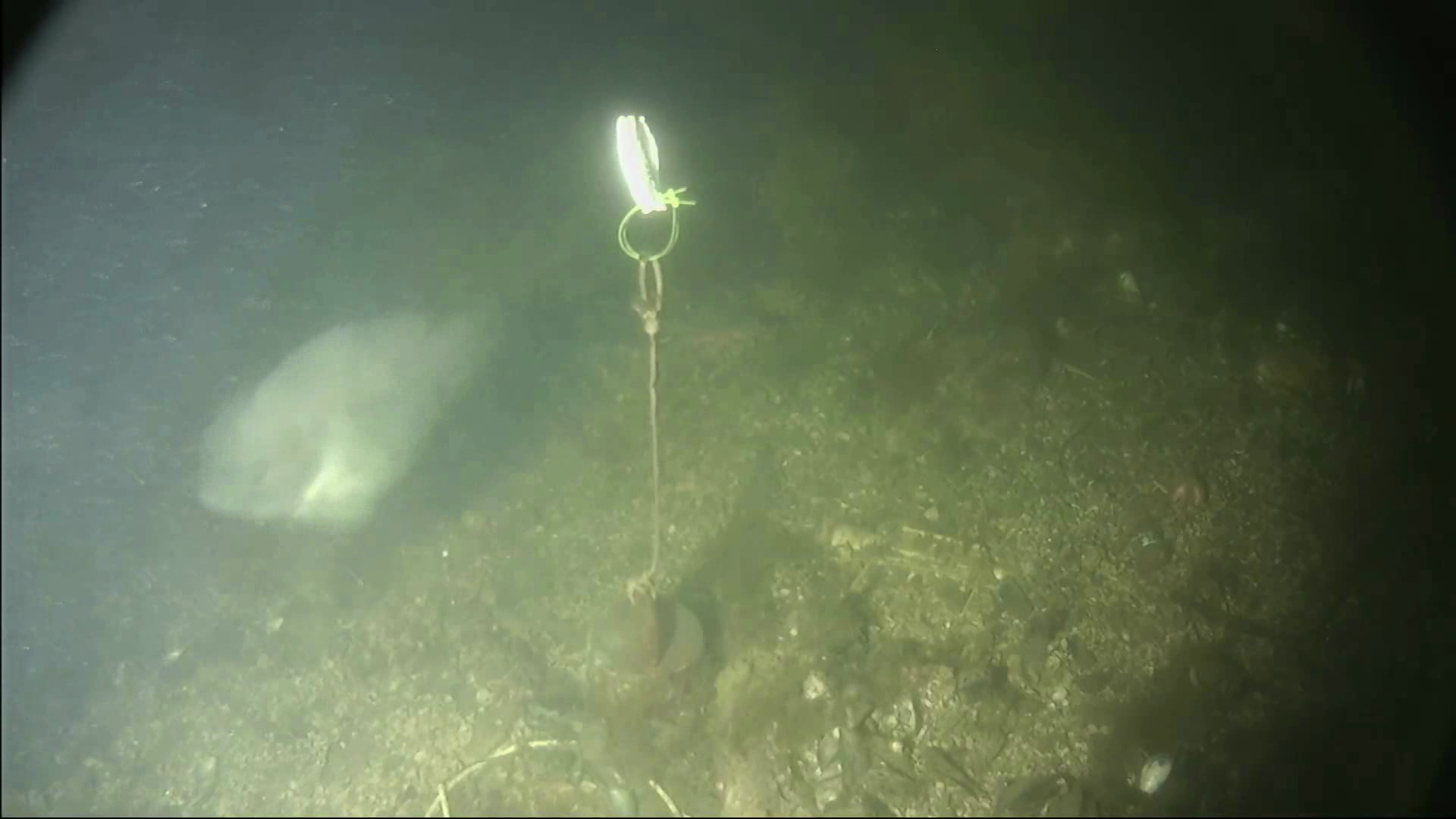}

    \caption{Sample images of anomalies in scene A (top) and B (bottom) in AURA: Anomalous Underwater Reef Activity.}
    \label{fig:var-visibility}
\end{figure*}

\subsection{Marine Life Datasets}
A wide variety of datasets includes video or image data of marine life, but they are focused on other tasks than VAD, such as object detection~\cite{Jansen2022FishDataset, pedersen2019detection}, classification~\cite{farrell2023labeled} or segmentation~\cite{saleh2020realistic, sauder2025coralscapesdatasetsemanticscene}.
This means that marine life is typically present in all frames
~\cite{Jansen2022FishDataset, sauder2025coralscapesdatasetsemanticscene}, making them infeasible for an investigation of VAD methods.
Additionally, some datasets rely on non-static cameras attached to divers or UAVs~\cite{sauder2025coralscapesdatasetsemanticscene}.
These datasets are also disregarded, as it is deemed infeasible for VAD methods to function properly in such scenarios. 


Possible options in terms of datasets suitable with enough empty frames was found to be the NOAA Puget Sound Nearshore Fish dataset~\cite{farrell2023labeled}, the Brackish dataset~\cite{pedersen2019detection} and subsets of the DeepFish dataset~\cite{saleh2020realistic}.

Another challenge besides identifying suitable datasets is also that they need to be re-annotated for the task of VAD in terms of marine life monitoring.
Namely, the existing annotations may not align with the goal of extracting interesting or anomalous events, as this is likely subjective and may vary from person to person.
The underwater setting complicates this task even further compared to other on-land datasets. Factors such as water turbidity, marine snow and varying light conditions cause substantial variation in visibility and image quality. 
This challenge will be addressed by using multiple annotators, as done in similar cases where the underwater settings could contribute to uncertainty in the annotations~\cite{HumblotRenaux2025Jambo}.

\section{The AURA Dataset}
In this section, we describe the content of the proposed AURA: Anomalous UnderwateR Activity dataset. 
The AURA dataset contains data from two locations, denoted scene A and scene B.
Samples from both locations can be seen in \cref{fig:var-visibility} and each scene is described in greater details in the following sections.

\subsection{Scene A (Anemo)}
\begin{figure}[t]
\centering
    \includegraphics[width=0.7\linewidth]{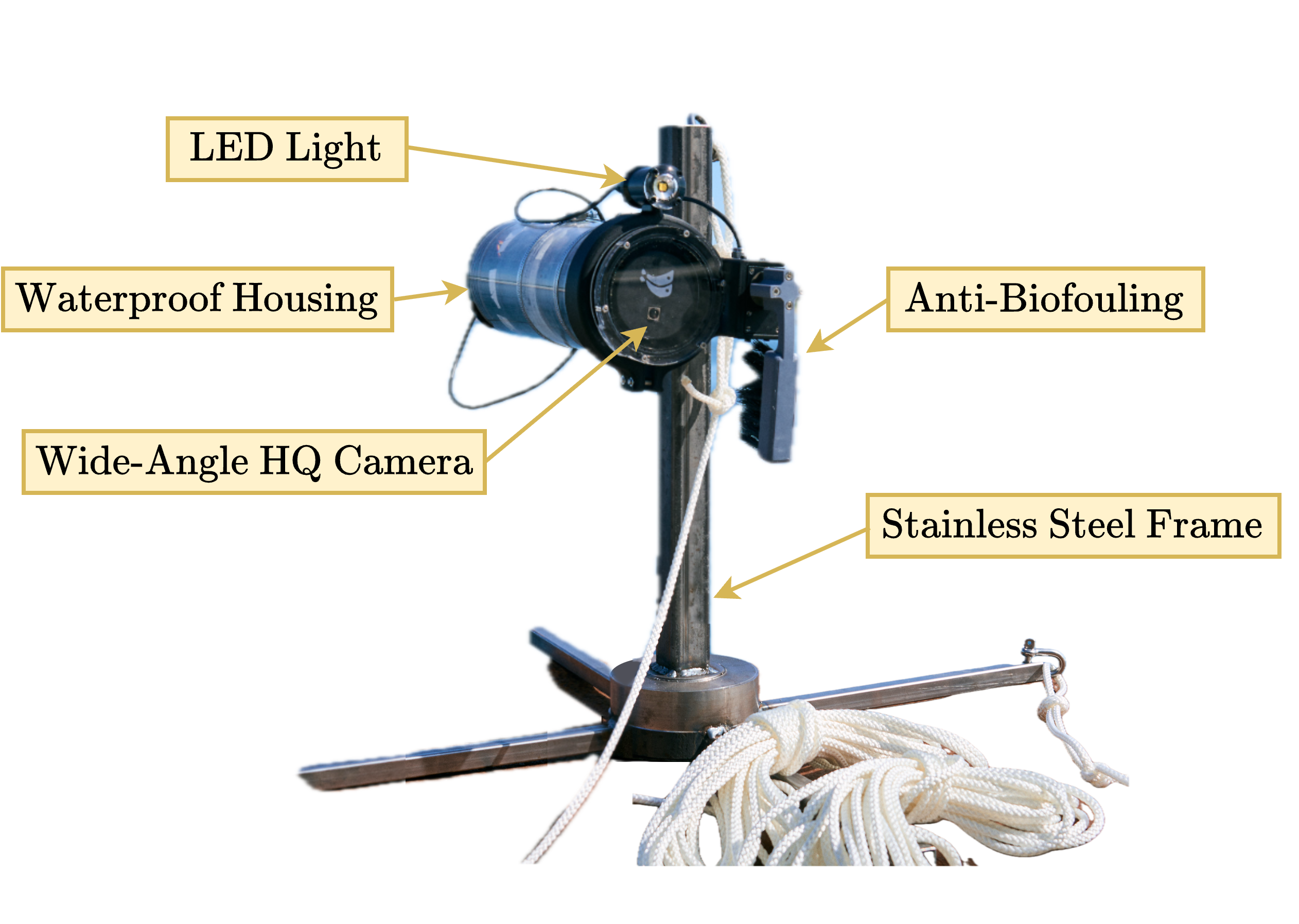}
    \caption{The AnemoCam features an adjustable LED light and a wide-angle high-resolution camera. To avoid buildup of biofouling, a mechanical wiper periodically sweeps the camera lens.}
  \label{fig:anemocam}

\end{figure}

Between July 2024 and February 2025, we deployed AnemoCam (see \cref{fig:anemocam}), a long-term underwater camera system in Hundested Harbour, Denmark. 
The camera is statically mounted on a stainless-steel frame at 11m depth. The field of view is roughly divided into four parts, showing an artificial reef, a sandy bottom, a water column, and a harbor wall. 
The artificial reef is intended to attract marine fauna and enhance benthic biodiversity within the harbor. 
Power is supplied by an exchangeable lithium-ion battery pack, allowing the system to operate autonomously for up to 3 months. 
The AnemoCam was configured to record 60-second video clips every 30 minutes, 24 hours per day. 

\subsection{Scene B (Brackish)}
We also use a subset of videos from the Brackish dataset \cite{pedersen2019detection} to include data from a different scene and camera setup. The camera system was mounted on a pillar of the Limfjords-bridge in Denmark at approximately 9m depth, positioned above a protective boulder barrier that also serves as habitat for marine species. The field of view captures the seafloor environment, sediment, and surrounding benthic habitat which is lit by artificial light.

\subsection{Data Compilation}
A total of 25 videos were selected; 10 videos from scene A (12,524 frames) and 15 from scene B (2,559 frames). Each video was selected such that it is possible to thoroughly identify only one anomalous event such as a fish or crab moving in and out of the field of view. If there are multiple anomalous events present, the video was trimmed down for simplicity.
The videos were selected to cover different times of day, varying levels of visibility and marine snow, and types of biological activity.




\subsection{Data Annotation}
In order to annotate the anomalous underwater events in the dataset we need a clear definition of what is meant by this.
Inspired by similar terminology from action spotting in sports \cite{xu2025action} we define it as follows:

\begin{definition}: A Contextually Bounded AnomalouS Sequence (C-BASS) is a visually interpretable and temporally bounded event that exhibits anomalous visual characteristics relative to the surrounding footage.
Every C-BASS has a start and end frame, marking the boundaries of the event.
\end{definition}

A C-BASS might be a fish moving into the field of view of a recorded underwater scene and towards the camera. 
The C-BASS starts at the frame where the fish first enters the scene and ends directly after the fish has left the field of view completely. A C-BASS can only be defined in the context of a full video, since the same video might also contain other ``less interesting'' biological activity. For example, a crab might sit in a different part of the scene throughout the main C-BASS. Using the C-BASS definition, each video was annotated multiple times by different annotators.
Each annotator was given the exact same instructions in writing (see supplementary material) for consistency.


\subsubsection{AnomaTag}
\begin{figure}[!htb]
\centering
    \includegraphics[width=\linewidth]{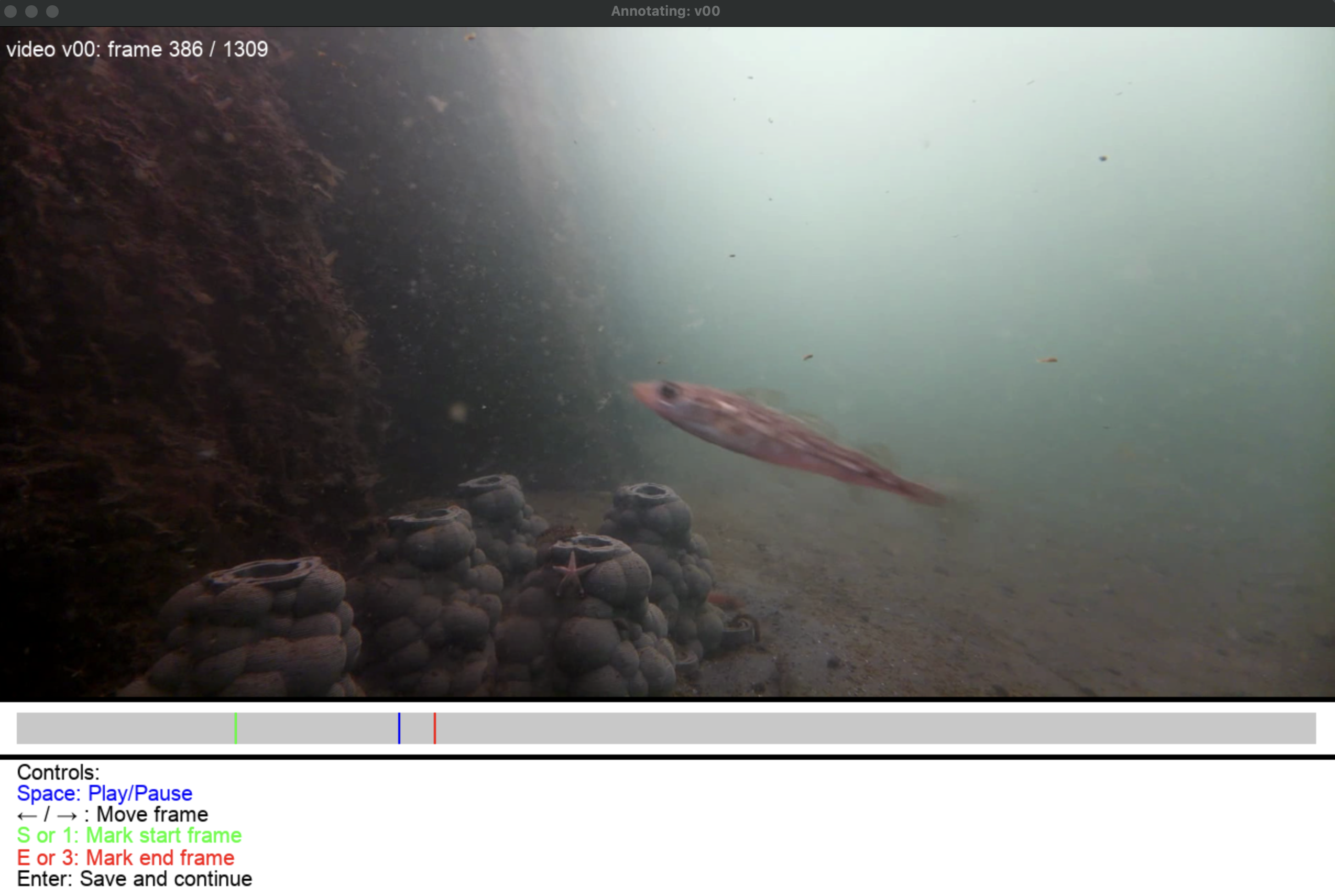}
    \caption{A screenshot of our custom tool AnomaTag for our anomalous event annotation.}
    \label{fig:anomatag-screenshot}
\end{figure}

To facilitate fast and intuitive annotation process, we developed a custom lightweight, frame-based annotation tool tailored specifically for anomaly detection workflows, called AnomaTag. The tool provides a minimal interface for navigating through video frames and selecting two key markers per segment: start and end frames. Playback, timeline visualization, keyboard shortcuts, and immediate visual and audio feedback enable efficient annotation even for long video sequences. A screenshot of AnomaTag can be seen in \cref{fig:anomatag-screenshot}. Annotations are saved in a simple text format, making integration with downstream processing pipelines straightforward. While general-purpose annotation platforms like Label Studio \cite{labelstudio} support video data annotation, they are primarily designed for frame-level object detection and pixel-wise segmentation tasks, and often require complex configuration and backend services. These platforms lack simple mechanisms for selecting sparse frames in a user-friendly manner, which makes them suboptimal for use cases such as event-based anomaly detection, where only a few significant frames need to be marked quickly and reliably.
\looseness -1


\subsection{Dataset Analysis}

A total of 16 persons annotated all of the 25 video sequences. The group of annotators consisted of 11 males and 5 females in the age range of roughly 25--45.
It should be noted that most of the annotators have a background in computer science but not necessarily computer vision or machine learning.
The pair-wise agreement between all of the annotators was calculated on a frame-by-frame level, which is depicted in \cref{fig:annotator-agreement}. We generally observe moderate to good agreement and most values fall in the 0.4-0.8 range, indicating moderate to substantial agreement. 
The average agreement can be estimated to be around $\kappa = 0.6$, which can be considered ``good'' for this subjective annotation task. The highest values are around ${\sim}0.79$, showing marine life event annotation is inherently subjective. We can observe some outlier annotators in \emph{U11} and \emph{U15}.

\subsubsection{Soft Labels}
\begin{figure}[]
\centering
    \includegraphics[width=0.95\linewidth]{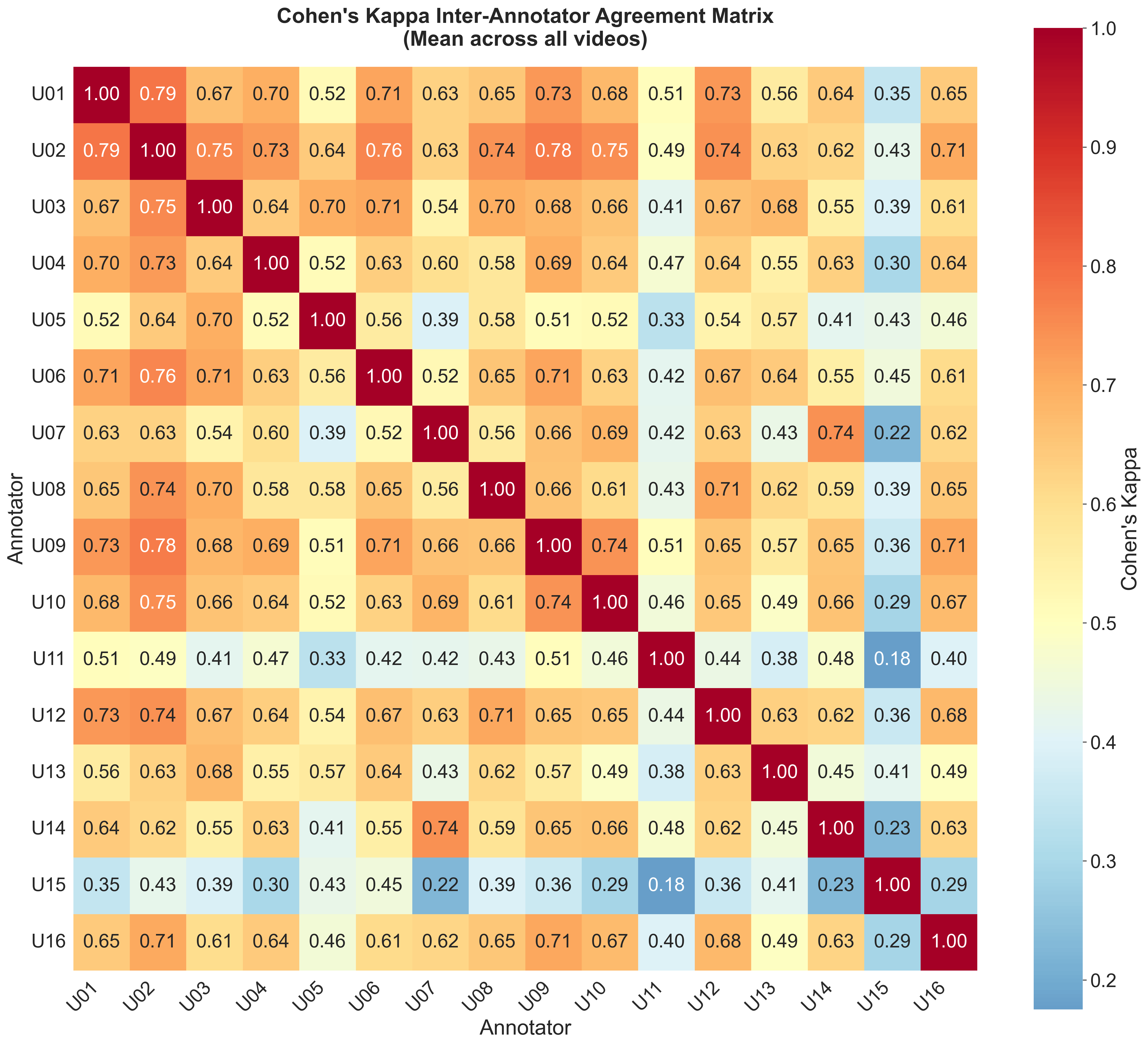}
    \caption{Cohen's Kappa scores between annotators.}
  \label{fig:annotator-agreement}
\end{figure}

The high variation in Cohen's Kappa scores seen in \cref{fig:annotator-agreement} motivates our usage of soft labels.
The soft labels are calculated for each video sequence, where the start and end frames from each annotator is converted into binary vectors of the same length as the video. These vectors are then averaged frame by frame across all annotators to produce soft labels that reflect the level of agreement over time by:


\begin{equation}
\bar{l}_{i,v} = \frac{1}{N} \sum_{a=1}^{N} l_{i,v}^{(a)}
\end{equation}

\noindent
where $\bar{l}_{i,v}$ is the resulting soft label for frame $i$ in video $v$ and $l_{i,v}^{(a)}$ is the binary label from annotator $a$.

\begin{figure*}[!htb]
\centering
    \includegraphics[width=0.9\linewidth]{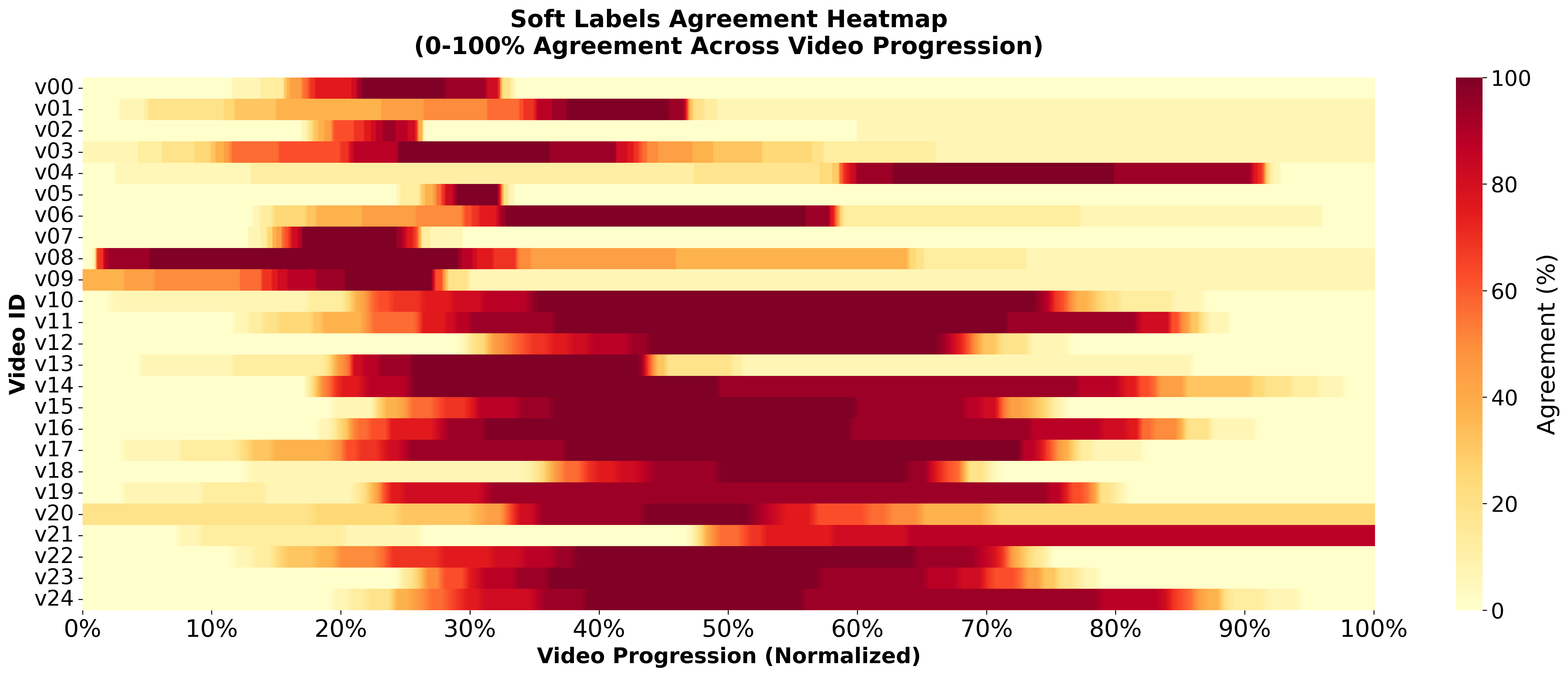}
    \caption{Temporal distribution of soft labels across all 25 videos in the AURA dataset. Each row represents one video sequence with frame numbers normalized for visualization. Color intensity indicates the proportion of annotators marking each frame as anomalous.}
  \label{fig:analysis-softlabel}
\end{figure*}

\begin{figure*}[!htb]
    \centering
    \begin{subcaptionbox}{Frame 230 - soft label $= 0.0$.\label{fig:softlabels-bio-activity-a}}[0.32\linewidth]
        {\includegraphics[width=0.95\linewidth]{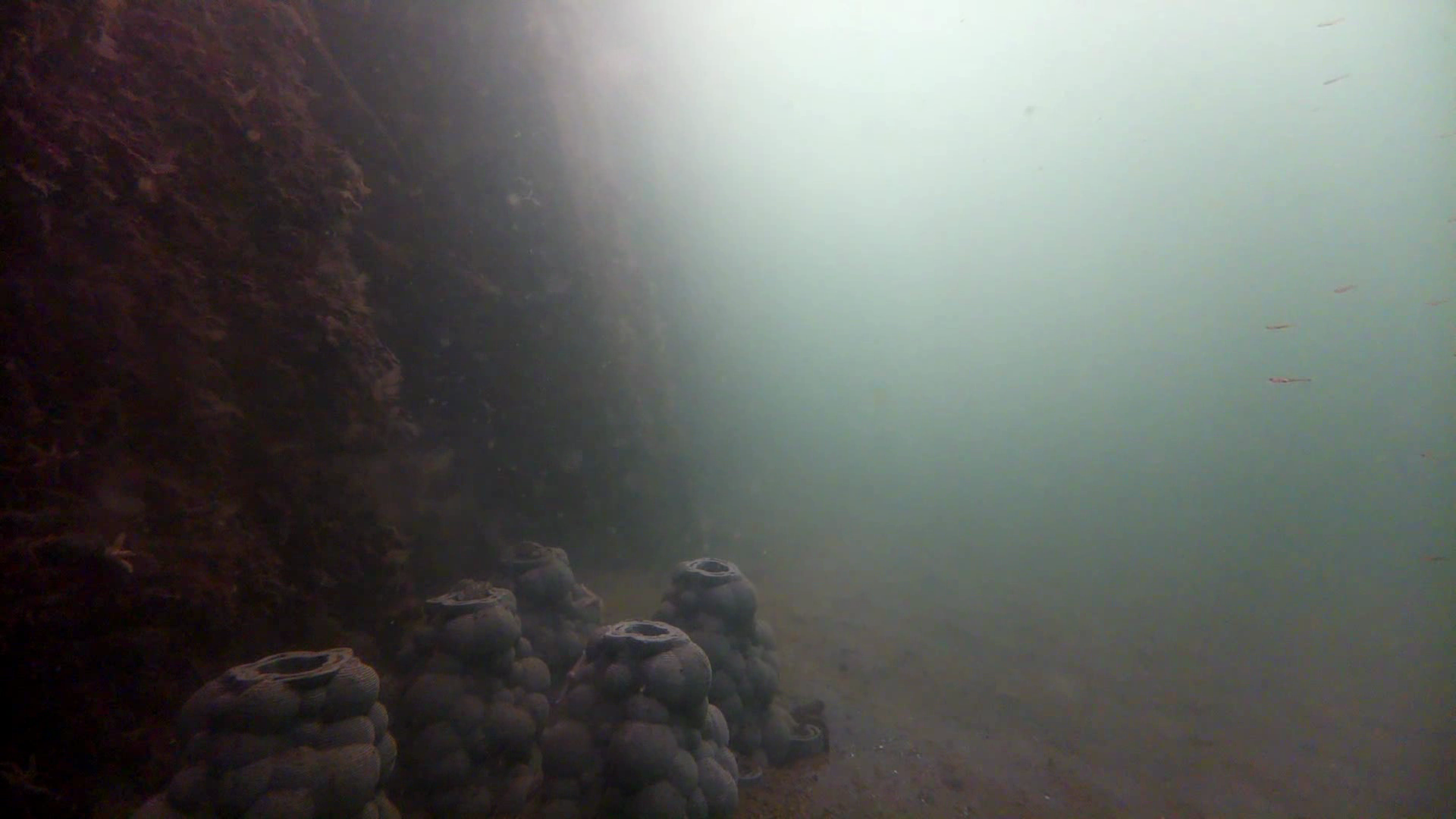}}
    \end{subcaptionbox}
    \begin{subcaptionbox}{Frame 370 - soft label $= 0.5$.\label{fig:softlabels-bio-activity-b}}[0.32\linewidth]
        {\includegraphics[width=0.95\linewidth]{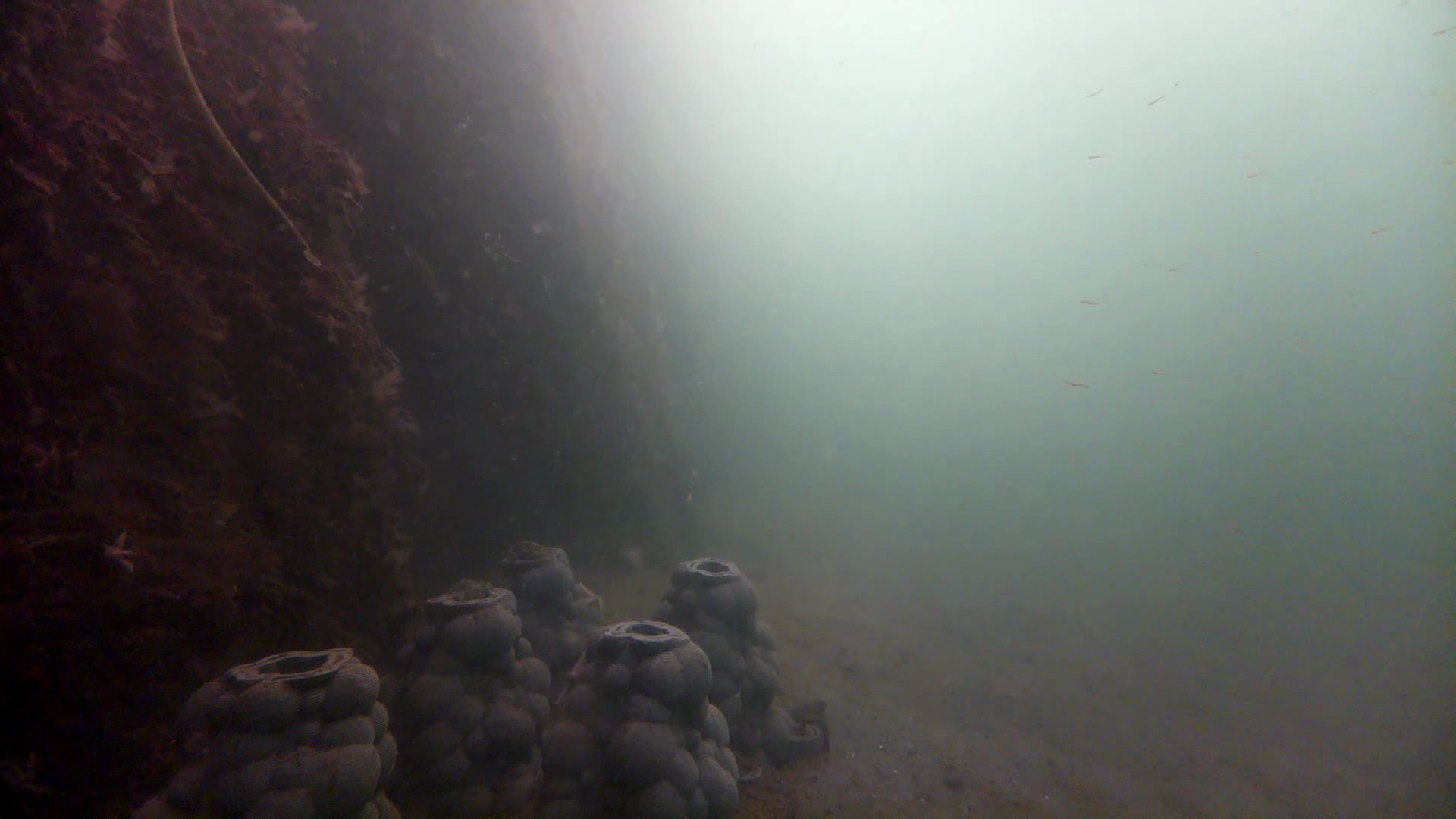}}
    \end{subcaptionbox}
    \begin{subcaptionbox}{Frame 510 - soft label $= 1.0$.\label{fig:softlabels-bio-activity-c}}[0.32\linewidth]
        {\includegraphics[width=0.95\linewidth]{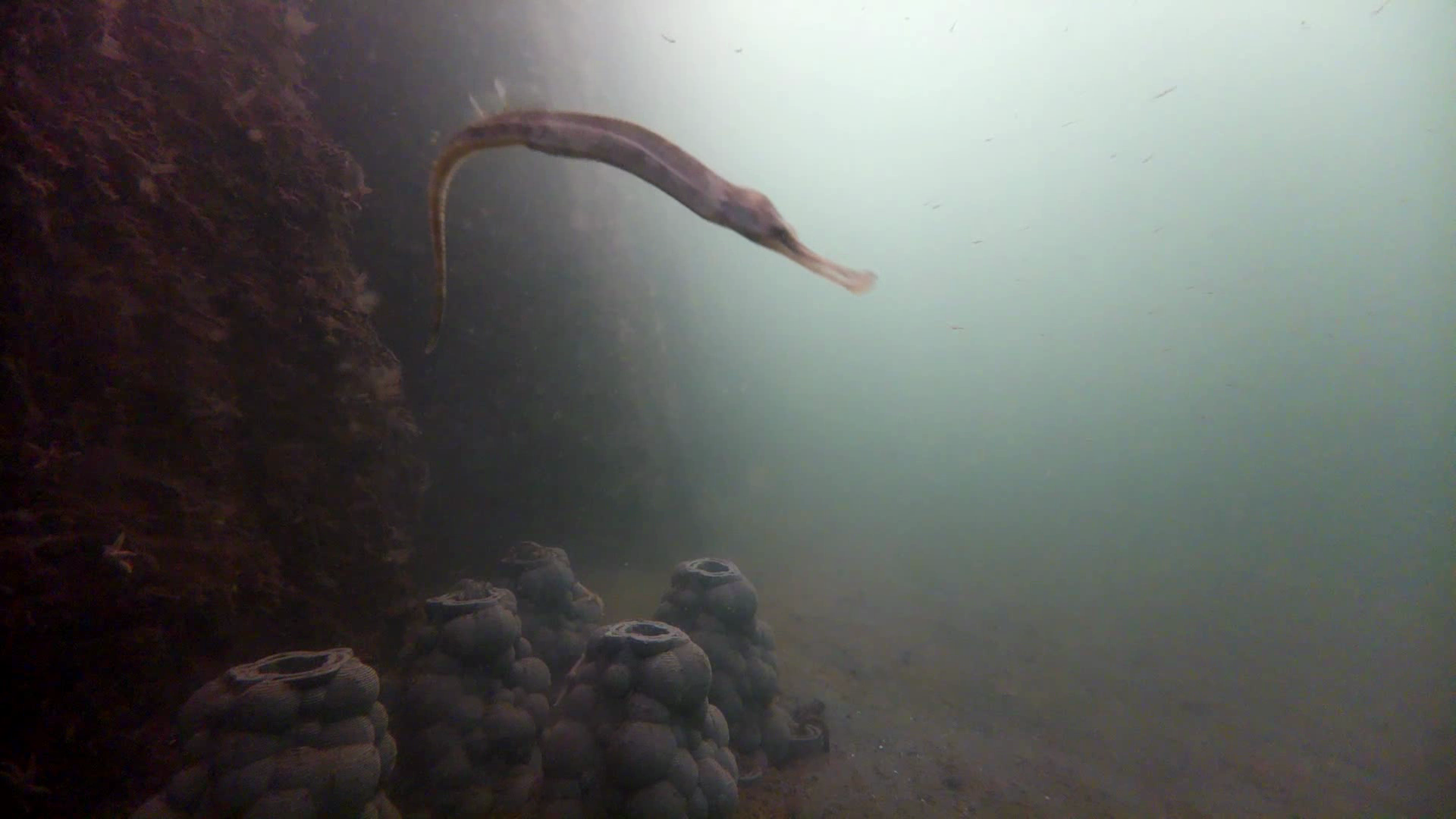}}
    \end{subcaptionbox}
    \caption{Frames from video \emph{v01} with varying levels of biological activity and the resulting soft labels. Frame (a) shows the empty scene. Only half of our annotators (soft label $= 0.5$) annotated the fish partially in view in the upper left corner in (b), whereas all of them annotated it later in (c).}
    \label{fig:softlabels-bio-activity}
\end{figure*}

A general overview of the distribution of the soft labels for all 25 videos can be seen in \cref{fig:analysis-softlabel}, where  each row represents a video sequence and the colormap indicates the soft labeling. 
The horizontal axis represents the frame number of each video. Note that the frame number has been normalized to account for the varying length of the videos for visualization purposes. 
It can be observed that videos in the AURA dataset generally have their peak soft label values centered in the middle of the sequence. 
This is especially true for the videos related to scene B (the last 15 videos) whereas the videos from scene A (the first 10 videos) are more diverse in the distribution.
In some cases, like video \emph{v05}, we can observe very distinct but brief events, whereas the soft label values change more gradually for other videos, such as \emph{v20}.
Furthermore, some videos, like \emph{v08}, appear to have the C-BASS start right at the beginning of the sequence whereas we can observe the opposite for video \emph{v21}. 
These soft labels aggregated from the different annotators are hence the closest it is possible to get to a real ground truth for the proposed dataset.
Some examples of soft labels for video $\emph{v01}$ can be seen in \cref{fig:softlabels-bio-activity}.
For \cref{fig:softlabels-bio-activity-a} and \cref{fig:softlabels-bio-activity-c}, all the annotators appear to agree on either the absence (label $= 0.0$) or presence (label $= 1.0$) of the pipefish.
However, for frame \cref{fig:softlabels-bio-activity-b} we observe a soft label of $0.5$ indicating that half of the 16 annotators observed the presence of the pipefish's tail in the upper left corner of the frame and deemed it interesting.
This supports the value of multiple annotators in this domain, as this observation might have been missed without it.

\subsubsection{Consensus Labels}
In addition to the soft labels, we also calculate the consensus labels, which are the integer frame numbers marking the start and end points of anomalous events in the videos. 
The consensus labels are therefore useful if we want to cut out interesting or anomalous image sequences, as we need integer indices for deciding where to cut the videos. Hence, we calculate the consensus labels by averaging the annotations across annotators for each temporal marker.
For $N$ annotators, the consensus start frame $\bar{s}_v$ and end frame $\bar{e}_v,$ are: 
\looseness -1

\begin{equation}
(\bar{s}_v,\; \bar{e}_v) = \frac{1}{N} \sum_{a=1}^{N} (s^{(a)}_v,\; e^{(a)}_v).
\end{equation}

\section{Visual Anomaly Detection}
In the following, we describe the overall pipeline that we employ for the VAD on the proposed AURA dataset, along with the different models that are evaluated and how they are trained. 
We also touch upon the issue of converting continuous anomaly scores into binary labels, which is important for the final evaluation of the different VAD methods.


\subsection{VAD Pipeline}

Our anomaly detection pipeline is based on deep neural networks (DNNs) and follows the flow shown in \cref{fig:intro-figure}. 
In this paper we solely focus on frame-based approaches because they are agnostic to video-length and FPS, providing better flexibility for real-world deployments where recording setups can vary.
First, we extract ``normal'' frames from the videos for the DNN training process, such that the model learns what the scene typically looks like. 
The trained model will, generally speaking, result in a low error for normal images as it was optimized for this during the training process.
However, when presented with data outside the normal, the model is expected to predict higher errors. 
During inference, it is hence possible to use this error as an expression for the normal-ness of the input image, which is commonly expressed as an anomaly map or a single anomaly score, often derived from the anomaly map.
For simplicity, our pipeline focuses only on the final anomaly score predicted per image.
By feeding the pipeline an entire video, frame-by-frame, we obtain a time series of anomaly scores, as illustrated in \cref{fig:intro-figure}.
The final step of the pipeline involves interpreting these anomaly score signals to identify the anomalous event and thereby the C-BASS.

\subsection{Anomalous Frame Selection}

We consider the problem of converting the visual anomaly scores from the videos to a C-BASS, thus ultimately finding discrete frame indices for trimming such videos into highlight sequences. This objective might seem trivial but in practice can dramatically reduce labeling efforts of vast amounts of video data for long-term monitoring efforts.


\paragraph{Problem Statement.} Let a video $v =\{i_1, i_2,... i_T\}$ consist of a sequence of $T$ images, and let $f_\text{anomaly}(i_t) = r_t \in \mathbb{R}$ denote the anomaly scores for frame $i_t$, then the resulting sequence of scores is $R = [r_1, r_2, \ldots, r_T]$. 
Our goal is to select a single contiguous interval $[s, e]$ of a start frame $s$ and end frame $e$, where $1 \leq s < e \leq T$ that represents the most prominent anomalous event in the video $v$. 
For event-based anomaly detection, we need a function $f_\text{select}$ that, given a sequence of scores $R$, maps the sequence to a binary label $\{0, 1\}$ where $1$ means \textit{anomalous} and $0$ means \textit{normal}.
\looseness -1

We consider two methods for selecting this interval from the anomaly score signal: (1) a threshold-based method, and (2) a peak-based method. We want to compare these two frame selection methods and how well they agree with the consensus labels from the annotators. We make the assumption that the most anomalous event is also represented by the longest temporally anomalous sequence in the anomaly scores. This allows us to focus on the most suggested anomalous event while discarding shorter, potentially noisy detections.

\paragraph{Thresholding method.} We apply a naive threshold $\tau$ to our anomaly scores and subsequently identify the longest contiguous segment of 1s and retain only that segment as the predicted anomalous event, setting all other values to 0. For all scores $R = [r_1, r_2, \ldots, r_T]$ with $t \in T$, we have:

\begin{equation}
f_\text{threshold}(\tau, R) =
\begin{cases}
1 & \text{if } t \in \underset{[s, e] \subseteq \{t \mid r_t \geq \tau\}}{\arg\max} (e - s + 1) \\
0 & \text{otherwise}.
\end{cases}
\end{equation}

The issue of the thresholding-based method is that it is highly sensitive to the choice of $\tau$ which can be difficult to tune, as it depends heavily on the VAD model, the specific video, scene conditions, and the type of anomaly.


\paragraph{Peak-based method.} An alternative approach for frame selection uses peak detection to identify a single dominant peak in our score signal. 
Given the anomaly score sequence $R$, we apply the \verb|find_peaks| function from scipy \cite{scipy}, to identify local maxima that may correspond to anomalous events within the sequence as follows:

\begin{equation}
f_\text{find\_peaks}(h, R) =
\begin{cases}
1 & \text{if } t \in \max \left( f_\text{peak\_widths}(R, h) \right) \\
0 & \text{otherwise}.
\end{cases}
\end{equation}

For each detected peak, we estimate the peak width using the \verb|peak_widths| function from scipy at a relative height parameter $h$, yielding candidate intervals around each peak. 
Among all detected peaks, we select the widest one, i.e. the one with the largest width at height $h$, assuming that more prominent anomalies correspond to broader peaks. 
This gives us the event boundaries $s$ and $e$ of the predicted anomalous event. 


\section{Evaluation}

We choose four DNN anomaly detection models from \emph{anomalib} \cite{akcay2022anomalib} and compare their performance on event-based anomaly metrics on our new AURA dataset. 
The chosen models range from more recent state-of-the-art~\cite{batzner2024efficientad} to older approaches~\cite{akcay2018ganomaly} to cover a variety of methods.
We evaluate the following models:

\begin{itemize}
    \item Reverse Distillation \cite{deng2022anomaly} uses a student decoder to learn to reconstruct a teacher encoder’s features from a compact embedding trained on normal data, with reconstruction failures representing anomalies.
    \item GANomaly \cite{akcay2018ganomaly} uses an encoder-decoder-encoder GAN to compare latent representations of input and reconstructed images, where anomalies show large differences in the latent space.
    \item Student-Teacher Feature Pyramid Matching (Stfpm) \cite{wang2021student} matches multi-scale feature pyramids between a teacher and student network, and detects anomalies by measuring discrepancies across corresponding layers.
    \item EfficientAD \cite{batzner2024efficientad} uses a fast student–teacher model and an auxiliary autoencoder to detect structural and logical anomalies with low latency.
\end{itemize}

\subsection{Training \& Dataset Splits}

The training split was selected based on visual inspection of the videos, aiming to capture representative examples of the ``normal'' background scene for each location. For scene A, which generally exhibits higher visual variability in environmental factors, training videos were chosen based on good visibility, high brightness, and low levels of marine snow. This subjective filtering ensured the model was trained on representative conditions. In contrast, scene B shows more consistent visual conditions, so training selection was guided by ensuring a clear visual distinction between normal and anomalous events. This strategy promotes more robust learning of scene-specific normality for anomaly detection. The key motivation for these splits is that we aim to explore how well the models generalize under different scene characteristics. Specifically, the performance on scene A with more training data and higher visual variability, versus scene B with fewer training frames but more consistent conditions. We evaluate on all videos, but for scene A and scene B separately (see \cref{tab:training-splits}). 
For both scene A and B, we also introduce an additional split for each, where split 2 contains twice as many videos in the training data as in split 1.
The addition of splits 1 and 2 serves to evaluate the impact of providing the VAD models with more training data.

\begin{table}[htpb]
    \centering
    \caption{Training videos for each split. Videos were selected based on visibility, brightness, and anomaly clarity.}
    \label{tab:training-splits}
    \begin{tabular}{@{}llcl@{}}
        \textbf{Split} & \textbf{Scene} & \textbf{\# Images} & \textbf{Video IDs} \\
        \midrule
        \multirow[c]{2}{*}{Split 1} 
            & Scene A & 3387 & v02, v03, v06, v09 \\
            & Scene B & 508  & v10, v12, v13, v18, v20 \\
        \midrule
        \multirow[c]{3}{*}{Split 2} 
            & Scene A & 6516 & \makecell[l]{v01, v02, v03, v05,\\ v06, v08, v09} \\
            & Scene B & 844  & \makecell[l]{v10, v12, v13, v14, v15,\\ v18, v20, v21, v23, v24} \\
    \end{tabular}
\end{table}


\subsection{Results}

\begin{table*}[!htbp]
\centering
\caption{Mean absolute error ($\downarrow$): Model predictions vs Soft Labels by Scene and Train Split}
\label{tab:mean-absolute-errors}
\resizebox{0.81\linewidth}{!}{%
\begin{tabular}{llcccc}
&& \multicolumn{2}{c}{\textbf{Scene A} (v00-v09)} & \multicolumn{2}{c}{\textbf{Scene B} (v10-v24)} \\
\textbf{Split} & \textbf{Model} & \textbf{MAE (Mean ± Std)} & \textbf{Best Video} & \textbf{MAE (Mean ± Std)} & \textbf{Best Video} \\
\midrule

Split 1 & EfficientAD & 0.420 ± 0.071 & v03 (0.303) & 0.384 ± 0.049 & v20 (0.269) \\
Split 1 & GANomaly & 0.526 ± 0.139 & v03 (0.330) & 0.435 ± 0.080 & v20 (0.332) \\
Split 1 & Stfpm & 0.432 ± 0.078 & v03 (0.325) & 0.433 ± 0.089 & v20 (0.237) \\
Split 1 & Reverse Distillation & \textbf{0.354} ± 0.153 & \textbf{v10 (0.167)} & \textbf{0.316} ± 0.094 & \textbf{v10 (0.167)} \\
\midrule
Split 2 & EfficientAD & 0.393 ± 0.101 & v03 (0.275) & 0.379 ± 0.048 & v20 (0.270) \\
Split 2 &GANomaly & 0.425 ± 0.127 & v03 (0.320) & 0.435 ± 0.099 & v14 (0.338) \\
Split 2 &Stfpm & 0.465 ± 0.107 & v10 (0.216) & 0.320 ± 0.058 & v10 (0.216) \\
Split 2 &Reverse Distillation & \textbf{0.271} ± 0.151 & \textbf{v01 (0.121)} & \textbf{0.259} ± 0.041 & \textbf{v24 (0.193)} \\
\end{tabular}
}
\end{table*}

To quantify the performance of the DNN-based VAD methods, we report the mean absolute error (MAE) between the normalized anomaly scores and the soft labels averaged across all videos per scene in \cref{tab:mean-absolute-errors}. Reverse Distillation outperforms all models showing the lowest values for MAEs across both data splits and scenes. Overall, most models show lower MAEs for scene B than scene A. 
Comparing MAEs across split 1 and 2 also suggests that most of the evaluated approaches benefit from the larger and more diverse training split.


We evaluate the effectiveness of both proposed frame selection methods: the threshold-based method $f_\text{threshold}(\tau, R)$  and the peak-based method $f_\text{find\_peaks}(h, R)$
Since both approaches depend on a single tuneable parameter, we conduct a dense parameter sweep across both $\tau \in [0, 1]$ and $h \in [0, 1]$ in 100 uniform increments (i.e., step size of 0.01). 
For each parameter setting, we apply the selection function to extract the C-BASS from the model's anomaly score sequence for each video and compute the temporal intersection over union (t-IoU):

\begin{equation}
\textrm{t-IoU}(\hat{E}, E)=\frac{|\hat{E} \cap E|}{| \hat{E}  \cup E|},
\end{equation}

\noindent
where $\hat{E} = [\hat{s}, \hat{e}]$ is the predicted event and $E = [\bar{s}, \bar{e}]$ is the consensus label. We report the best average t-IoU across all videos for each scene and data split, i.e., the optimal performance achieved by each model–selection pair under its best-tuned parameter (see supplementary material for more details). These results are shown in \cref{tab:best-params-split}.

\begin{table}[htbp]
\centering
\caption{Best parameter and average temporal IoU ($\uparrow$) by frame selection method and scene.}
\label{tab:best-params-split}
\resizebox{\linewidth}{!}{%
\begin{tabular}{@{}llcccc@{}}
\multirow{2}{*}{\textbf{Model}} & \multirow{2}{*}{\textbf{Method}} & \multicolumn{2}{c}{\textbf{Scene A}} & \multicolumn{2}{c}{\textbf{Scene B}} \\
\cmidrule(lr){3-4} \cmidrule(lr){5-6}
& & \textbf{Param} & \textbf{t-IoU} & \textbf{Param} & \textbf{t-IoU} \\

\toprule
\multicolumn{6}{l}{\textbf{Split 1}} \\
\midrule
EfficientAD & Find Peaks & 0.98 & 0.451 & 1.00 & 0.557 \\
GANomaly & Find Peaks & 0.87 & 0.179 & 1.00 & 0.373 \\
Reverse Distillation & Find Peaks & 0.76 & \textbf{0.534} & 0.97 & \textbf{0.717} \\
Stfpm & Find Peaks & 0.89 & 0.447 & 0.82 & 0.429 \\
\midrule
EfficientAD & Threshold & 0.51 & 0.297 & 0.51 & 0.400 \\
GANomaly & Threshold & 0.79 & 0.120 & 0.60 & 0.163 \\
Reverse Distillation & Threshold & 0.67 & \textbf{0.453} & 0.72 & \textbf{0.513} \\
Stfpm & Threshold & 0.57 & 0.314 & 0.73 & 0.347 \\
\midrule
\multicolumn{6}{l}{\textbf{Split 2}} \\
\midrule
EfficientAD & Find Peaks & 1.00 & 0.347 & 0.99 & 0.745 \\
GANomaly & Find Peaks & 1.00 & 0.242 & 0.94 & 0.471 \\
Reverse Distillation & Find Peaks & 0.88 & \textbf{0.673} & 0.81 & 0.861 \\
Stfpm & Find Peaks & 0.86 & 0.301 & 0.97 & \textbf{0.863} \\
\midrule
EfficientAD & Threshold & 0.49 & 0.374 & 0.50 & 0.709 \\
GANomaly & Threshold & 0.46 & 0.083 & 0.80 & 0.217 \\
Reverse Distillation & Threshold & 0.37 & \textbf{0.461} & 0.61 & \textbf{0.806} \\
Stfpm & Threshold & 0.63 & 0.225 & 0.56 & 0.748 \\

\end{tabular}%
}
\end{table}

\begin{figure}[!htbp]
\centering
    \includegraphics[width=0.48\linewidth]{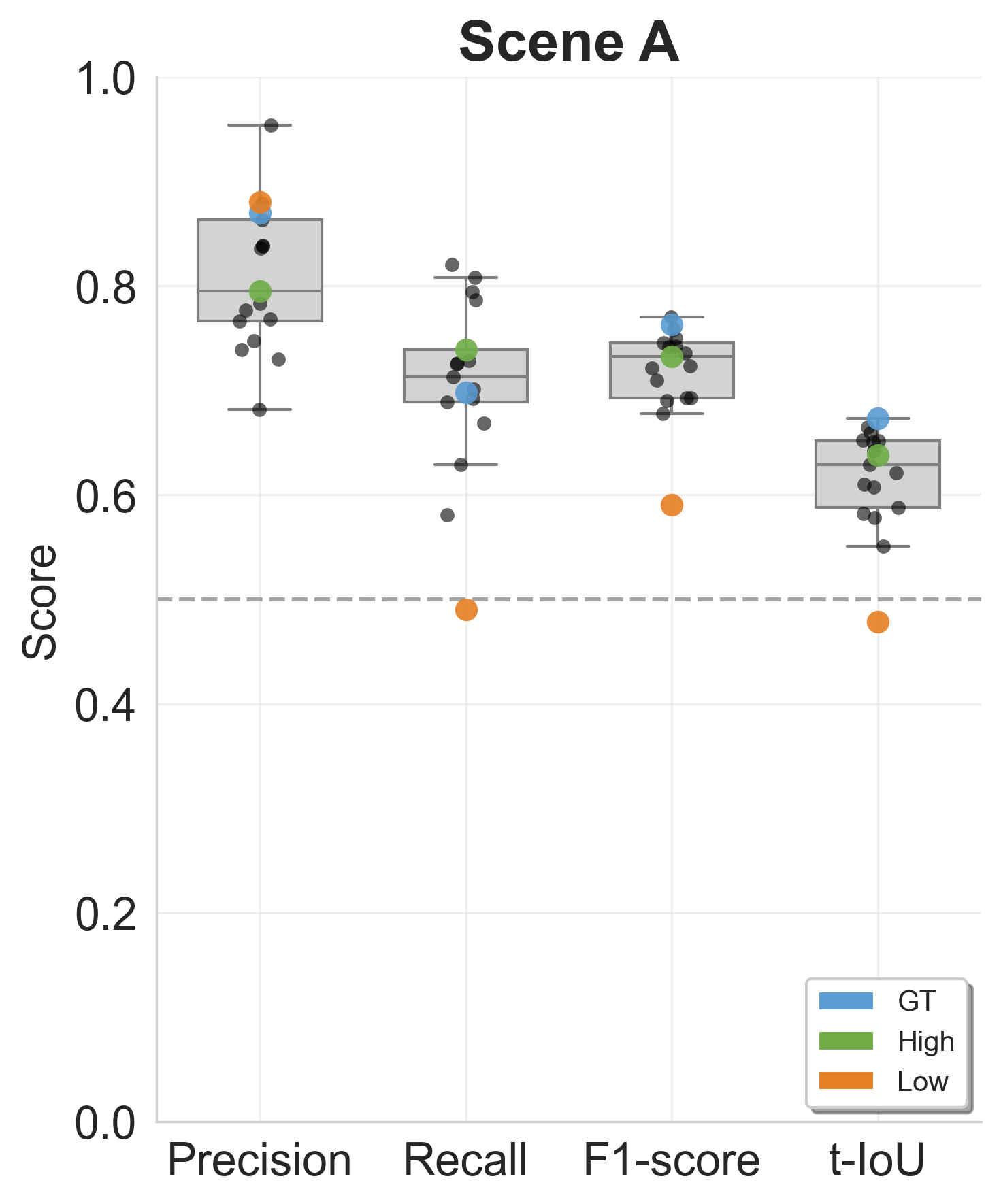}
    \includegraphics[width=0.48\linewidth]{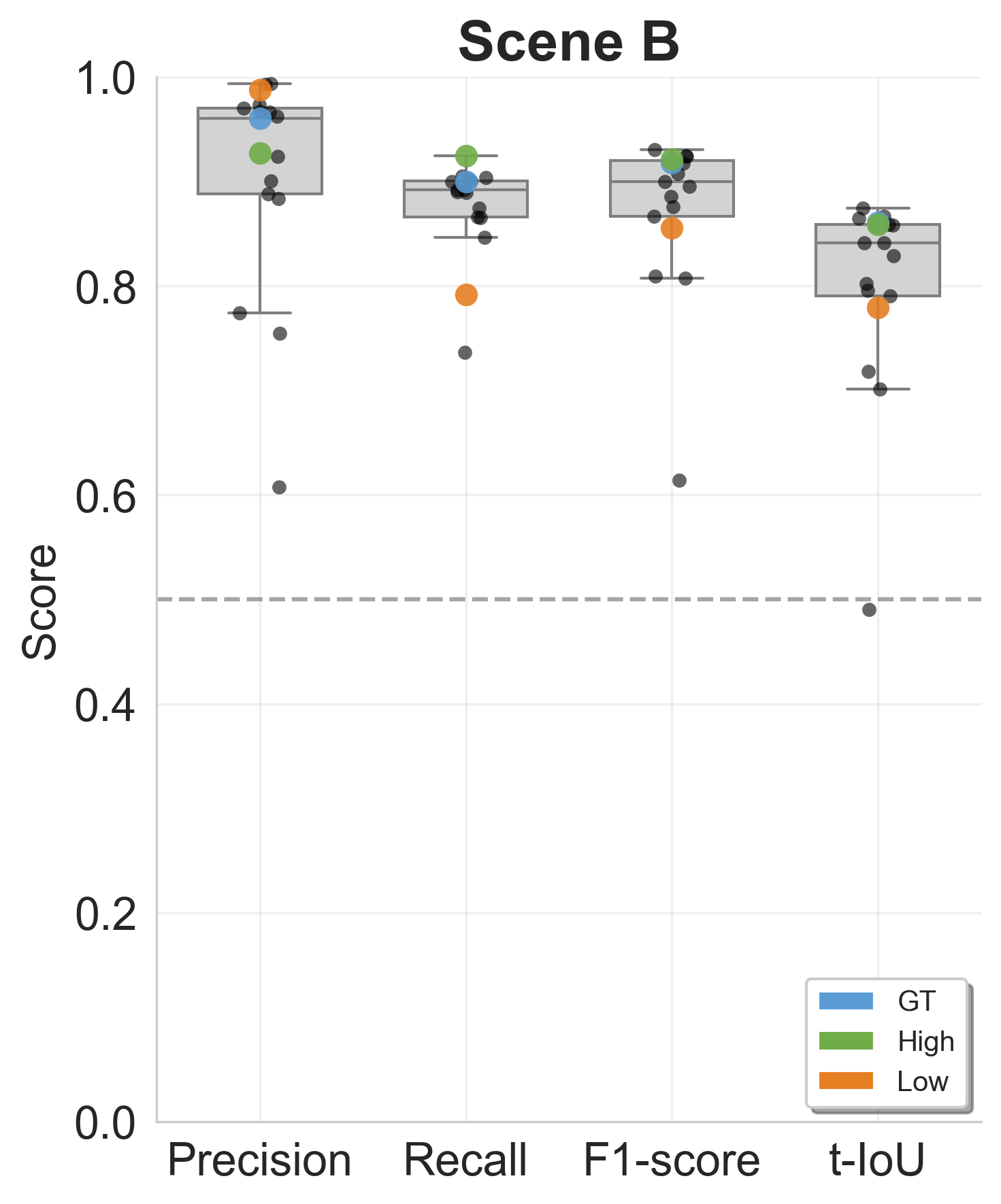}
    \caption{The distribution of precision, recall, F1, and t-IoU scores per scene, where each dot represents a single annotator.   
    The colors highlight the consensus labels (blue), a high-agreement \protect\emph{U02} (green), and a low-agreement annotator \protect\emph{U11} (orange) as per their Cohen's Kappa scores (see \cref{fig:annotator-agreement}).}
    
  \label{fig:all-metrics}
\end{figure}

We observe that Reverse Distillation performs best overall across both scene and data splits, indicating that it consistently generates anomaly scores that are temporally well-aligned with the human perceived anomalies in our data set. 
The frame selection method based on peak finding outperforms naive thresholding in nearly all models and scenes. 
This suggests peak-finding is a more robust approach for detecting C-BASS segments in our data set. 
Scene B shows higher values for t-IoU which aligns with our expectations, as its C-BASS segments tend to be more visually distinct.


To demonstrate the importance of having multiple annotators in the AURA dataset, we calculate the precision, recall, F1 score and t-IoU according to the labels made by each annotator and plot the resulting distribution in \cref{fig:all-metrics}.
For this evaluation, we use the best best-performing approach identified earlier, consisting of ReverseDistillation with peak-finding and the parameters identified in \cref{tab:best-params-split}.
Note that no re-training is done in this test and the same VAD model is used for all annotators.
The distribution of the different performance metrics across the annotators highlights, that the same model and parameter settings yield different results depending on which annotator’s labels are used as ground truth. 
Therefore, the evaluation is sensitive to annotator variability. 
These differences suggest that an annotator can over- or underestimate the length and position of events, supporting the need for multiple annotators.



\section{Conclusion}
We introduce AURA, the first visual anomaly detection benchmark in underwater video data for biodiversity monitoring with multi-annotator labels. 
We evaluate four VAD models, demonstrating their capability to detect anomalous events in underwater environments.
Among the evaluated methods, Reverse Distillation consistently showed the best alignment with human annotations.
Additionally, we find that peak-based selection methods are more effective than naive thresholding across frame-by-frame anomaly scores in video sequences. 
Lastly, we conclude that model performance is sensitive to variability in ground truth labels, marking the importance of soft and consensus labels. 
Overall, our findings suggest that VAD models, when paired with multi-annotator labels and frame selection techniques, offer a promising path toward scalable, camera-agnostic marine monitoring systems.
Future work could include the expansion of the AURA dataset with data from other locations and with longer sequences, which could contain multiple anomalous events of interest.

\section*{Acknowledgments}
This project has received funding from the European Union’s Horizon 2020 research and innovation programme under the Marie Skłodowska-Curie grant agreement No. 956200. Additionally, the project has received funding as part of AI Denmark, financed by The Danish Industry Foundation.

{
    \small
    \bibliographystyle{ieeenat_fullname}
    \bibliography{lib/misc, lib/vad, lib/fish, lib/cv}
}
\appendix

\section{Supplementary Material}

\subsection{Labeling Instructions} \label{app:annotation-procedure}
We show our labeling instructions for the annotation task of anomalous events with a start and end frame in \cref{{fig:anomatag-instructions}}.

\begin{figure}[!htbp]
\centering
    \includegraphics[width=\linewidth]{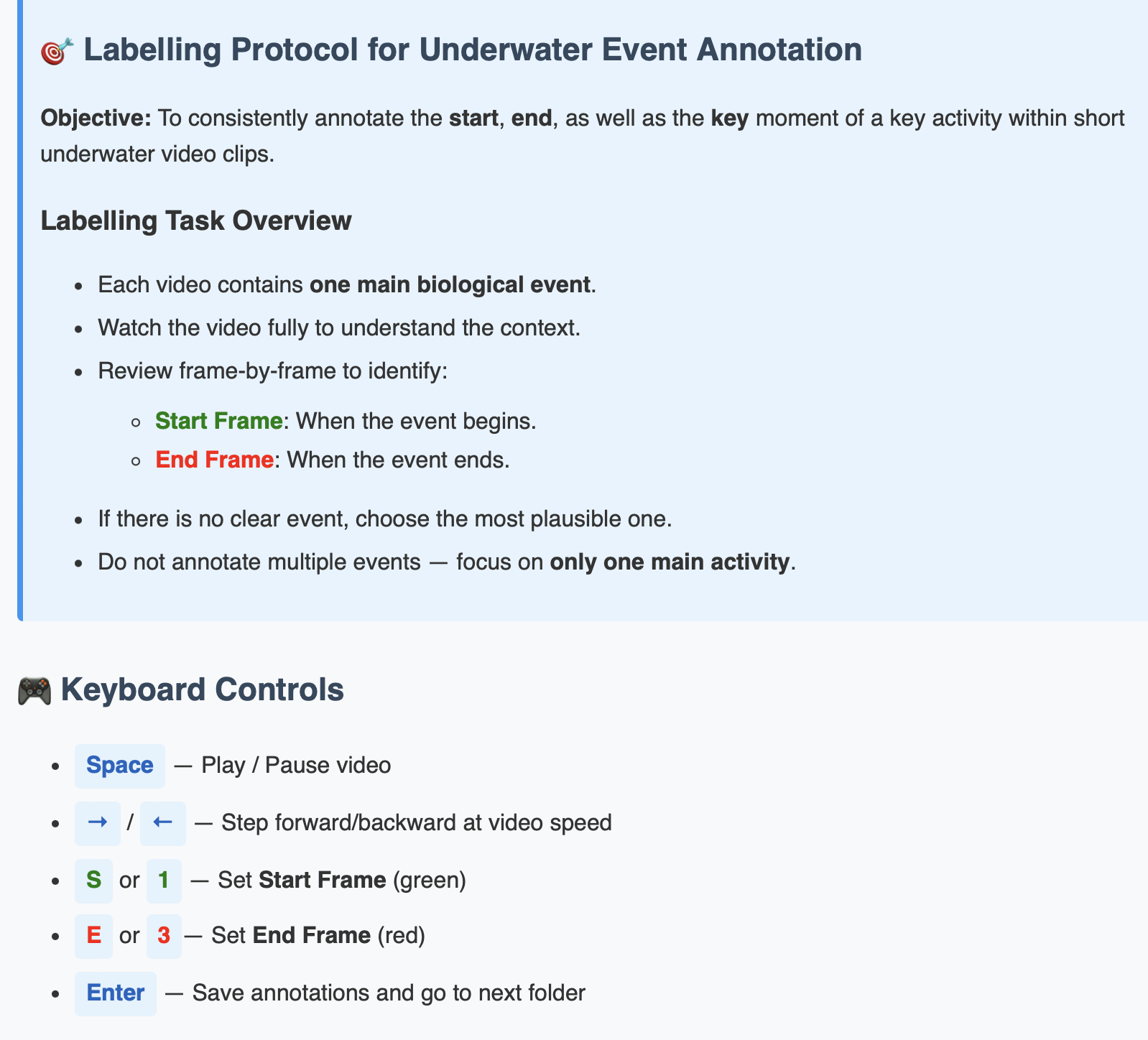}
    \caption{A screenshot of the labeling instructions for our annotation task in AnomaTag.}
  \label{fig:anomatag-instructions}

\end{figure}


\subsection{Parameter Sweep}
\begin{figure}
\centering
    \includegraphics[width=\linewidth]{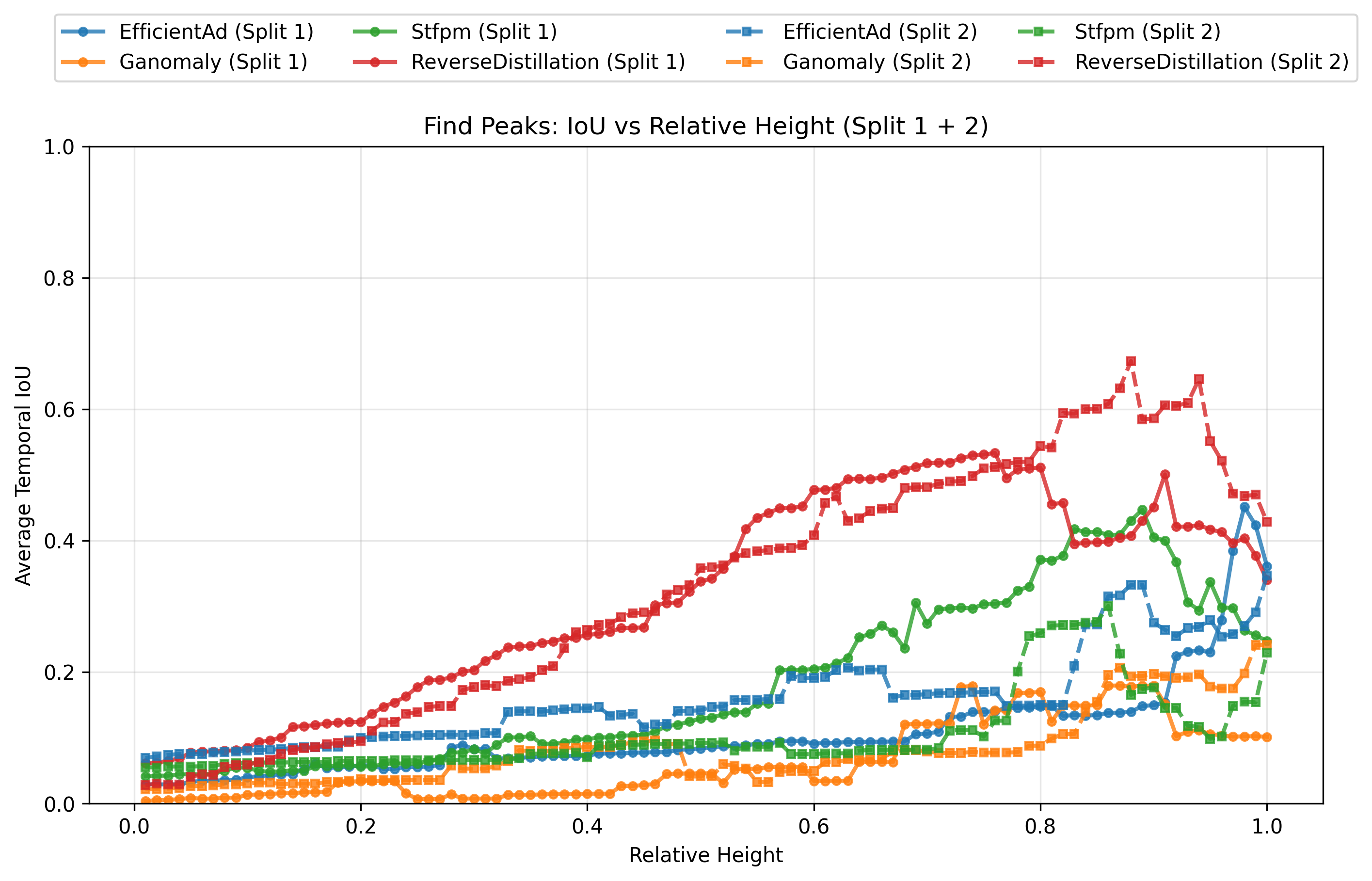}
    \includegraphics[width=\linewidth]{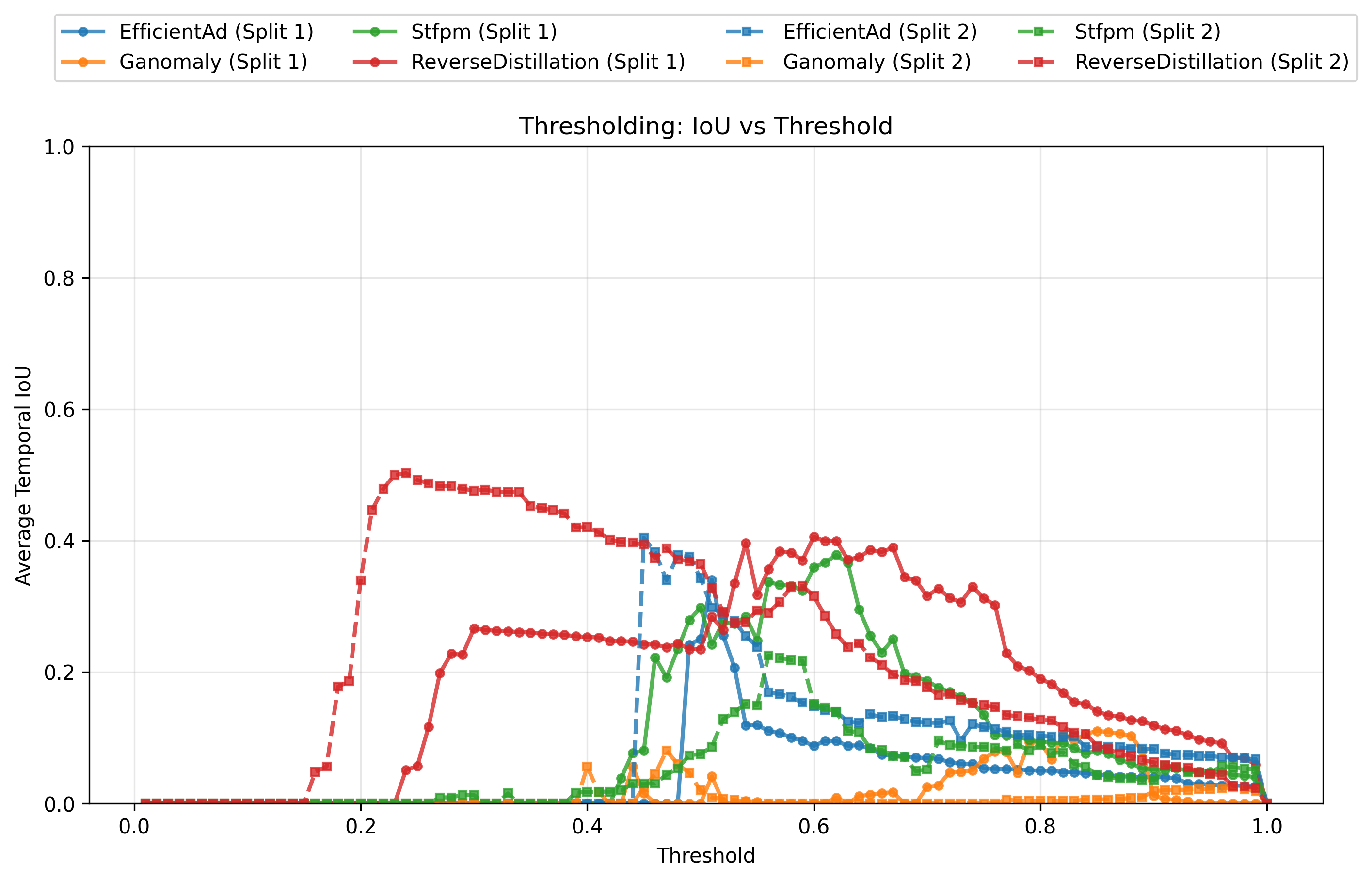}
    \caption{Average temporal IoU performance across parameter settings for  peak-finding (top) and thresholding methods (bottom) per data split. Performance is averaged across videos for each parameter value.}
  \label{fig:best-param-search}

\end{figure}

We show average temporal IoU performance across all parameter settings, which is \textit{relative height} for find-peak and $\tau$ for thresholding in \cref{fig:best-param-search}.

\end{document}